\documentclass[11pt,a4paper]{article}
\usepackage[hyperref]{acl2021}
\usepackage{times}
\usepackage{latexsym}

\usepackage{microtype}

\usepackage{amsmath}
\usepackage{xcolor}
\usepackage{array}

\usepackage{graphicx}
\usepackage{algorithm}
\usepackage{algorithmic}
\usepackage{booktabs}
\usepackage{multirow}
\usepackage{amsfonts}
\usepackage{amssymb}

\usepackage{nicefrac}
\usepackage{url}

\newcommand{\R}{\mathbb{R}}

\makeatletter
\newcommand{\ssymbol}[1]{$^{\@fnsymbol{#1}}$}
\makeatother

\newlength\savewidth
\newcommand\shline{\noalign{\global\savewidth\arrayrulewidth
                            \global\arrayrulewidth 0.8pt}
                   \hline
                   \noalign{\global\arrayrulewidth\savewidth}}

\aclfinalcopy

\title{Contrastive Attention for Automatic Chest X-ray Report Generation}

\author{Fenglin Liu\textsuperscript{1}, Changchang Yin\textsuperscript{3}, Xian Wu\textsuperscript{4},  Shen Ge\textsuperscript{4}, Ping Zhang\textsuperscript{3}, Yuexian Zou\textsuperscript{1}, and Xu Sun\textsuperscript{2}\\
\textsuperscript{1}ADSPLAB, School of ECE, Peking University\\
\textsuperscript{2}MOE Key Laboratory of Computational Linguistics, School of EECS, Peking University\\
\textsuperscript{3}The Ohio State University \ \ \textsuperscript{4}Tencent Medical AI Lab, Beijing, China\\
{\tt fenglinliu98@pku.edu.cn; \{kevinxwu, shenge\}@tencent.com} \\ 
{\tt \{yin.731, zhang.10631\}@osu.edu; \{zouyx, xusun\}@pku.edu.cn}\\ 
}

\date{}

\begin{document}
\maketitle
\begin{abstract}
Recently, chest X-ray report generation, which aims to automatically generate descriptions of given chest X-ray images, has received growing research interests. The key challenge of chest X-ray report generation is to accurately capture and describe the abnormal regions. In most cases, the normal regions dominate the entire chest X-ray image, and the corresponding descriptions of these normal regions dominate the final report. Due to such data bias, learning-based models may fail to attend to abnormal regions. In this work, to effectively capture and describe abnormal regions, we propose the Contrastive Attention (CA) model. Instead of solely focusing on the current input image, the CA model compares the current input image with normal images to distill the contrastive information. The acquired contrastive information can better represent the visual features of abnormal regions. According to the experiments on the public IU-X-ray and MIMIC-CXR datasets, incorporating our CA into several existing models can boost their performance across most metrics. In addition, according to the analysis, the CA model can help existing models better attend to the abnormal regions and provide more accurate descriptions which are crucial for an interpretable diagnosis. Specifically, we achieve the state-of-the-art results on the two public datasets.

\end{abstract}

\begin{figure}[t]

\centering
\includegraphics[width=1\linewidth]{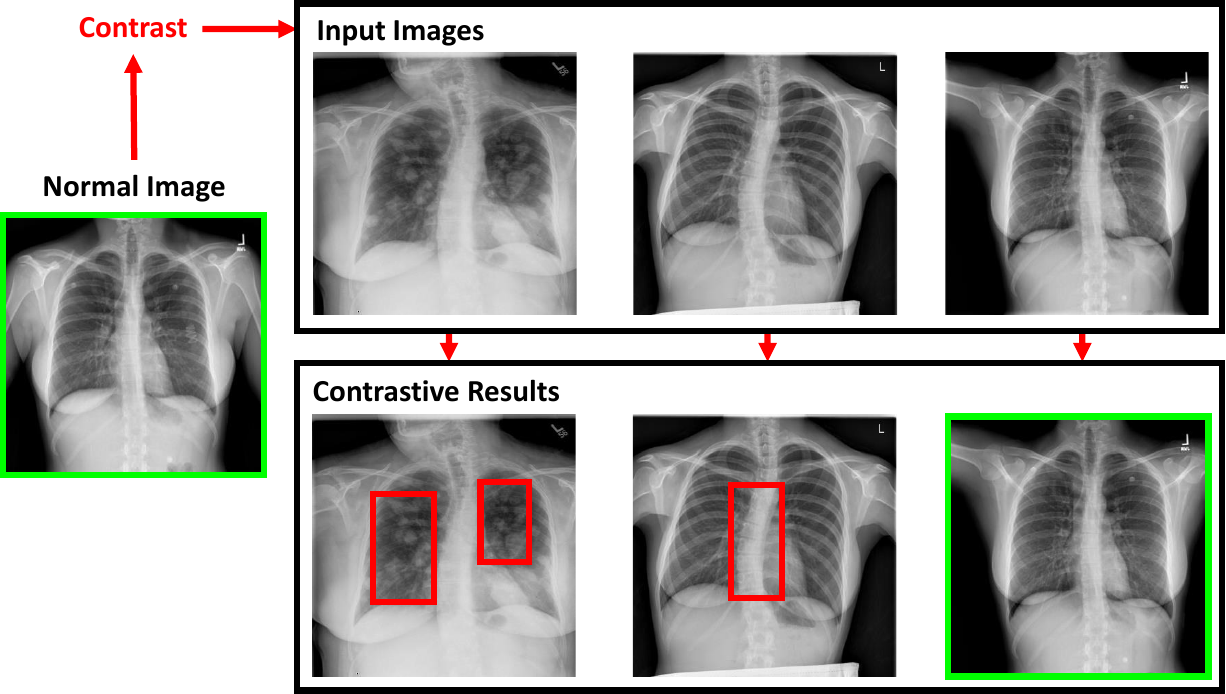}
\caption{By contrasting current input images and known normal images, it could be easier to capture the suspicious abnormal regions (Red bounding boxes). The images with Green boxes are normal.}
\label{fig:contrastive}
\end{figure}

\section{Introduction}
A medical report is a paragraph containing multiple sentences that describe both the normal and abnormal regions in the chest X-ray image. 
Chest X-ray images and their corresponding reports are widely used in clinical diagnosis \cite{Delrue2011Difficulties}. 
However, writing medical reports requires particular domain knowledge \cite{goergen2013evidence}, and only experienced radiologists can accurately interpret chest X-ray images and note down corresponding findings in a coherent manner.
An automatic chest X-ray report generation system \cite{Jing2018Automatic,Jing2019Show,fenglin2021PPKED,Liu2019Clinically} can reduce the workload of radiologists and are in urgent need \cite{Brady2012DiscrepancyAE,Delrue2011Difficulties}.

In recent years, several deep learning-based methods have been proposed \cite{Jing2018Automatic,Jing2019Show,Li2018Hybrid,Li2019Knowledge,Chen2020Generating,fenglin2021PPKED} for automatic chest X-ray report generation; however, there are serious data deviation problems in the medical report corpus. For example, 1) the normal images dominate the dataset over the abnormal ones \cite{Shin2016Learning}; 2) given an input image, the normal regions usually dominate the image and their descriptions dominate the medical report \cite{Jing2019Show,fenglin2021PPKED}. Such data deviation may prevent learning-based methods from capturing the rare but important abnormal regions (e.g., lesion regions). As a result, the learning-based model tends to generate plausible general reports with no prominent abnormal narratives \cite{Jing2019Show,Li2018Hybrid,Yuan2019Enrichment}. In clinical practice, 
accurate detection and depiction of abnormalities are more helpful in disease diagnosing and treatment planning.
Therefore, existing learning-based methods may fail to assist radiologists in clinical decision-making \cite{goergen2013evidence}.

\begin{figure*}[t]

\centering
\includegraphics[width=0.88\linewidth]{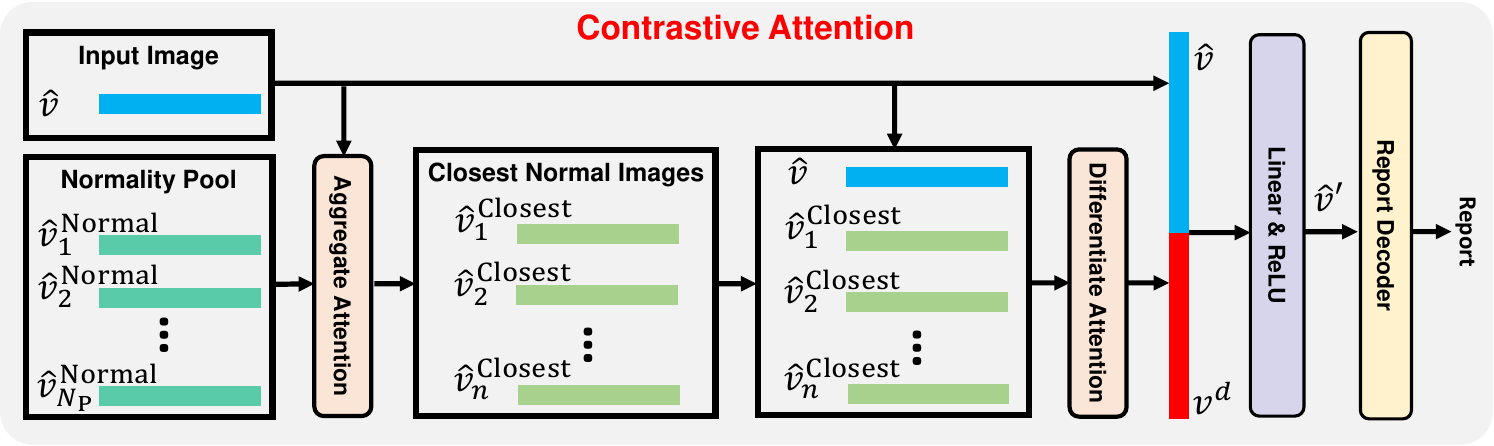}
\caption{Illustration of our proposed Contrastive Attention, which consists of the Aggregate Attention and Differentiate Attention. In particular, the Aggregate Attention devotes to finding the normal images that are closest to the current input image in the normality pool. The Differentiate Attention devotes to summarizing the common information between the input image and the closest normal images and subtract it from the input image to capture the differentiating properties between the input image and the normal images.}
\label{fig:model}
\end{figure*}

To capture the abnormal regions from a chest X-ray image, a natural intuition is to compare it with normal images and identify the differences. As Figure~\ref{fig:contrastive} shows, given known normal images, it cloud be easier for models to learn and identify the suspicious abnormal regions (Red bounding boxes). Therefore, we propose the Contrastive Attention (CA) model (see Figure~\ref{fig:model}), which is based on the attention mechanism \cite{bahdanau2014neural,Vaswani2017Transformer}. The CA model can be easily integrated into existing learning-based methods and enable them to better capture and describe the abnormalities. We build the CA model in the following three steps: 1) we first build a set of normal images which are all extracted from the training dataset; 2) we introduce the Aggregate Attention to prioritize normal images that are closer to the current input image, and filter out normal images that appear differently; 3) and we further introduce the Differentiate Attention to distill the common features between the input image and the refined normal images. Then, the acquired common features are subtracted from the visual features of the input image. In this manner, the residual visual features of the input image are 
treated as the contrastive information that captures the differentiating properties between input image and normal images.

We evaluate our approach on two datasets, including a widely-used benchmark IU-X-ray dataset \cite{Dina2016IU-Xray} and a recently released large-scale MIMIC-CXR dataset \cite{Johnson2019MIMIC}. On both automatic metrics and human evaluations, existing methods equipped with the proposed Contrastive Attention model outperform baselines, which proves our argument and demonstrates the effectiveness of our approach.

Overall, the main contributions of our work are:
\begin{itemize}

    \item We propose the Contrastive Attention model to capture and depict the abnormalities by comparing the input image with known normal images. The proposed approach can be easily incorporated into existing models to improve their performance.

    \item We evaluate our approach on two public datasets. After equipping our Contrastive Attention model, the baselines achieve up to 14\% gain and 17\% gain in BLEU-4 on the MIMIC-CXR and IU-X-ray datasets, respectively.
    
    \item More encouragingly, we achieve the state-of-the-art performance on the two public datasets, i.e., IU-X-ray and MIMIC-CXR. Moreover, we invite professional clinicians to conduct human evaluation to measure the effectiveness in terms of its usefulness for clinical practice.
    
\end{itemize}

\section{Related Works}
\label{sec:related_work}

\paragraph{Image Captioning}
Image captioning aims to understand the given images and generate corresponding descriptive sentences \cite{chen2015microsoft}.
The task combines image understanding and language generation.
In recent years, a large number of encoder-decoder based neural systems have been proposed for image captioning \cite{Cornia2020M2,Pan2020XLinear,Pei2019Memory,Venugopalan2015VC,Vinyals2015Show,Xu2015Show,rennie2017self,lu2017knowing,anderson2018bottom,fenglin2018simnet,fenglin2019GLIED}.
However, the sentence generated by image captioning is usually short and describes the most prominent visual contents, which cannot fully represent the rich feature information of the image.
Recently, visual paragraph generation \cite{Krause2017Hierarchical}, which aims to generate long and coherent reports or stories to describe visual contents, has recently attracted increasing research interests. 
However, due to the data bias in the medical domain, the widely-used hierarchical LSTM in the visual paragraph generation does not perform very well in automatic chest X-ray report generation and is tend to produce normal reports \cite{Xue2018Multimodal,Li2018Hybrid,Jing2019Show,Yin2019Automatic}.

\paragraph{Chest X-ray Report Generation}

Inspired by the success of deep learning models on image captioning, a lot of encoder-decoder based frameworks have been proposed \cite{Jing2018Automatic,Jing2019Show,fenglin2021PPKED,Liu2019Clinically,Yuan2019Enrichment,Xue2018Multimodal,Li2018Hybrid,Li2019Knowledge,Li2020Auxiliary,Zhang2020When,Kurisinkel2021Coherent,Ni2020Learning,Nishino2020Reinforcement,Chen2020Generating,Wang2021Unifying,Boag2019Baselines,Tanveer2020Label,Yang2020Automatic,Lovelace2020Learning,Zhang2020Optimizing,Miura2021Factual}.
Specifically, \citet{Jing2018Automatic} proposed a hierarchical LSTM with the attention mechanism \cite{Bahdanau2015NMT,Xu2015Show,You2016Image}. \citet{Yuan2019Enrichment} further incorporated the medical concept to enrich the decoder with descriptive semantics. \citet{Xue2018Multimodal} proposed a multimodal recurrent model containing an iterative decoder with visual attention to improve the coherence between sentences.
\citet{Miura2021Factual} proposed an Exact Entity Match Reward and an Entailing Entity Match Reward to improve the factual completeness and consistency of the generated reports, resulting in significant improvements on clinical accuracy.
\citet{Jing2019Show,Li2018Hybrid,Liu2019Clinically} and \citet{Zhang2020When,Li2019Knowledge,fenglin2021PPKED} introduced the reinforcement learning and medical knowledge graph for chest X-ray report generation, respectively.
However, some errors occur in the generated reports of the existing methods, like duplicate reports, inexact descriptions, etc \cite{Xue2018Multimodal,Yuan2019Enrichment,Yin2019Automatic}.

\paragraph{Contrastive Learning}
The most related work to our contrastive attention mechanism is in the field of contrastive learning \cite{Chen2020CL,He2020MoCo,Olivier2020CL,Grill2020Bootstrap,Chen2020MoCov2,Radford2021CLIP,Jia2021Scaling}, which learns similar/dissimilar image representations from data that are organized into similar/dissimilar image pairs.
In image captioning, \citet{Dai2017CL} introduced the contrastive learning to extract the contrastive information from additional images into the captioning models to improve the distinctiveness of the generated captions.
Moreover, \citet{Song2018CA} and \citet{Duan2019CA} proposed the contrastive attention mechanism for person re-identification and summarization, respectively. 
\citet{Song2018CA} utilized a pre-provided person and background segmentation to learn features contrastively from the body and background regions, resulting they can be easily discriminated.
\citet{Duan2019CA} contrastively attended to relevant parts and irrelevant parts of source sentence for abstractive sentence summarization.
In this work, we leverage the contrastive information between the input image and the normal images to help models efficiently capture and describe the abnormalities for automatic chest X-ray report generation.

\section{Approach}
\label{sec:approach}
In Section~\ref{sec:problem}, we formulate the automatic chest X-ray report generation problems; In Section~\ref{sec:contrastive}, we introduce the Contrastive Attention in detail.

\subsection{Problem Formulation}
\label{sec:problem}
Given a chest X-ray image $I$, the goal of automatic chest X-ray report generation is to generate a coherent report $R$ that addresses findings of $I$.
Most existing methods \cite{Jing2018Automatic} adopt the encoder-decoder frameworks, which normally include an image encoder \cite{he2016deep} and a report decoder \cite{Krause2017Hierarchical}. The encoder-decoder framework can be formulated as:
\begin{equation}
\label{eqn:formula}
\footnotesize
\text{Image Encoder}  : I \to V ;
\quad
\text{Report Decoder} : V \to R .
\end{equation}

The image encoder, e.g., ResNet \cite{he2016deep}, aims to generate the visual features $V \in \R^{{N_\text{I}} \times d}$. 
The report decoder, e.g., Hierarchical LSTM \cite{Krause2017Hierarchical}, is used to generate the report $R$ from $V$.
Specifically, in the Hierarchical LSTM, a paragraph-level LSTM first generates topic vectors to represent the sentences, then a sentence-level LSTM takes each topic vector as input to generate the corresponding sentence.
As a result, the Hierarchical LSTM can better model a paragraph of multiple sentences (report) than a single LSTM \cite{Jing2018Automatic,Krause2017Hierarchical}.
Given the ground truth medical report provided by the radiologists for the input chest X-ray image, existing methods train the encoder-decoder frameworks by minimizing training loss, e.g., cross-entropy loss.
Due to limited space, please refer to \citet{Huang2019Multi_Attention,Jing2018Automatic} for detailed introduction.

In this paper, we adopt the ResNet-50 \cite{he2016deep} to extract the visual features, i.e., the output of the last convolutional layer is used as the visual information:
\begin{equation}
\footnotesize
    V = \text{ResNet}(I) \text{W}^I ,  
\end{equation}
where $I$ denotes the input image, $\text{ResNet}(I) \in \R^{49 \times 2048}$ and $\text{W}^I \in \R^{2048 \times d}$ which reduces the dimension from 2,048 to $d$\footnote{For conciseness, all the bias terms of linear transformations in this paper are omitted.}.
Specifically, the $d$ is set to 512, resulting $V = \{v_1, v_2, \dots, v_{N_\text{I}}\} \in \R^{N_\text{I} \times d}$, where $N_\text{I}=49$ and $d=512$.
Moreover, we apply the average pooling to obtain the global visual feature:
\begin{equation}
\footnotesize
   \hat{v} = \text{AveragePooling}(V) = \frac{1}{{N_\text{I}}}\sum_{i=1}^{{N_\text{I}}} v_i .
\end{equation}

After the above calculations, we obtain the visual feature vectors $V = \{v_1, v_2, \dots, v_{N_\text{I}}\} \in \R^{N_\text{I} \times d}$ and the global visual feature vector $\hat{v} \in \R^{1 \times d}$.

\subsection{Contrastive Attention}
\label{sec:contrastive}
In this paper, we propose the Contrastive Attention to enable the models to capture the differentiating properties between the input image and normal images.
To this end, we first collect a normality pool $P = \{\hat{v}^\text{Normal}_1, \hat{v}^\text{Normal}_2, \dots, \hat{v}^\text{Normal}_{N_\text{P}}\} \in \R^{N_\text{P} \times d}$ which consists of $N_\text{P}=1,000$ normal images randomly extracted from the training dataset, where $\hat{v}^\text{Normal}_i \in \R^{1 \times d}$ denotes the global visual feature of $i^\text{th}$ extracted normal image.
Then, as shown in Figure~\ref{fig:model}, the proposed Contrastive Attention introduces the Aggregate Attention and Differentiate Attention which aims to obtain the contrastive information between the input image $\hat{v}$ and the normality pool $P$.

\paragraph{Aggregate Attention}
Since the images in the normality pool $P$ are all normal, there is no ranking order among these images. Therefore it's natural to treat all normal images equally in capturing the contrastive information. However, as shown in Figure~\ref{fig:example}, we note that there are many noisy images in the normality pool (see the Purple boxes in Figure~\ref{fig:example}). For example, some normal images have different orientation information or rotation angles from the input image, we cannot direct compare these images with the input image.
Therefore, these noisy images will bring noise information, preventing the Contrastive Attention from capturing accurate abnormal regions efficiently.
Motivated by the above observations, we introduce Aggregate Attention to increase the weights of normal images that are close to the current input image and lower the weights of images that are not close to the current input image.
As a result, the contrasting process could be improved.
To implement the Aggregate Attention, we utilize the dot-product attention\footnote{Our preliminary experiments show that using the dot-product attention \cite{Vaswani2017Transformer} could achieve better performance than the additive attention \cite{bahdanau2014neural,Xu2015Show,lu2017knowing} and bilinear attention \cite{Kim2018Bilinear,You2016Image}.} \cite{Vaswani2017Transformer}, which is defined as:
\begin{equation}
\footnotesize
\label{eqn:mha}
\begin{aligned}
\text{Att}(x,y)& =\text{softmax}(M) y \\
\text{where} \ \ M &= \frac{x \text{W}^x (y \text{W}^y)^T}{\sqrt{d}}  , 
\end{aligned}
\end{equation}
where $\text{W}^x, \text{W}^y \in \R^{d \times d}$ are learnable parameters. Given $x \in \R^{N_x \times d}$ and $y \in \R^{N_y \times d}$, the acquired $M$ is then of the shape of $N_x \times N_y$, the function $softmax$ is conducted on each row of $M$, resulting in $\text{Att}(x,y) \in \R^{N_x \times d}$. Then we apply Eq.~(\ref{eqn:mha}) to $\hat{v} \in \R^{1 \times d}$ and $P \in \R^{N_\text{P} \times d}$:
\begin{equation}
\footnotesize
\begin{aligned}
\label{eqn:dot-product}
\hat{v}^\text{Closest} &= \text{Att}(\hat{v},P) .
\end{aligned}
\end{equation}
Since the attention mechanism is a function that computes the similarity of $\hat{v}$ and each $\hat{v}^\text{Normal}_i$ in $P$, and $\text{Att}(\hat{v},P) \in \R^{1 \times d}$ is the attended vector for the $\hat{v}$.
In this way, we can increase the weights of normal images that are similar to current image and lower the weights of the ones that are dissimilar.

Moreover, following \citet{Lin2017Self_Attentive}, we repeat the Aggregate Attention $n$ times with different learnable attention weights to further promote the performance of our approach, defined as follows:
\begin{equation}
\label{eqn:n}
\footnotesize
\begin{aligned}
P' &= \text{Aggregate-Attention}(\hat{v}, P) \\
&= [\text{Att}_1(\hat{v},P); \text{Att}_2(\hat{v},P); \dots; \text{Att}_n(\hat{v},P)] \\
&= \{\hat{v}^\text{Closest}_1, \hat{v}^\text{Closest}_2, \dots, \hat{v}^\text{Closest}_n\} \in \R^{n \times d} ,
\end{aligned}
\end{equation}
where $[\cdot;\cdot]$ stands for concatenation operation.
In particular, for chest X-ray images, the result $\text{Att}(\hat{v},P)$ in Eq.~(\ref{eqn:dot-product}) turns out to be the found normal images that are closest to the current entire input image.
However, $\text{Att}(\hat{v},P)$ cannot capture those normal images that are closest to the input image only in a specific part rather than the entire image.
Fortunately, through repeating the function $\text{Att}(\hat{v},P)$ in Eq.~(\ref{eqn:dot-product}) $n$ times, our Aggregate Attention can efficiently find the closest normal images $P' = \{\hat{v}^\text{Closest}_1, \hat{v}^\text{Closest}_2, \dots, \hat{v}^\text{Closest}_n\}$ from $n$ parts.

\paragraph{Differentiate Attention} \
To learn the contrastive information between the input image and the closest normal images, we first attempt to capture their similarity, i.e., the common information, then we remove this similarity portion from the input image to obtain the contrastive information.
To this end, we introduce the Differentiate Attention.

The first step is learning to summarize the common information $v^c \in \R^{1 \times d}$ between the information of current input image $\hat{v} \in \R^{1 \times d}$ and the information of closest normal images $P' \in \R^{n \times d}$. 
In implementation, we employ the same dot-product self-attention mechanism in Eq.~(\ref{eqn:dot-product}) and average pooling operation to obtain the $v^c \in \R^{1 \times d}$:
\begin{align}
\footnotesize
v^c = \text{AveragePooling}\left(\text{Att}\left([\hat{v}; P'], [\hat{v}; P']\right)\right) ,
\end{align}
where $[\cdot;\cdot]$ denotes row-wise concatenation operation, then the $[\hat{v}; P']$ is in the shape of $(n+1) \times d$ and the $\text{Att}\left([\hat{v}; P'], [\hat{v}; P']\right)$ function outputs a matrix in the shape of $(n+1) \times d$.

In this way, via such self-attention mechanism, we exploit the similarity between $P'$ and $\hat{v}$ to capture the significant common information.
Next, to obtain the contrastive information $v^d \in \R^{1 \times d}$, we remove (i.e., subtract) the common information $v^c \in \R^{1 \times d}$ from the input image $\hat{v} \in \R^{1 \times d}$:
\begin{equation}
v^d = \hat{v} - v^c .
\end{equation}
At last, we update the original image features, i.e., $\hat{v} \in \R^{1 \times d}$ and $V = \{v_1, v_2, \dots, v_{N_\text{I}}\} \in \R^{{N_\text{I}} \times d}$, with the contrastive information $v^d \in \R^{1 \times d}$:
\begin{align}
\footnotesize
\hat{v}' &= \text{ReLU}([\hat{v}; v^d]\text{W}') \\ 
v'_i &= \text{ReLU}([v_i; v^d]\text{W}') ,
\end{align}
where $\text{ReLU}(\cdot)$ represents the ReLU activation function and $\text{W}' \in \R^{2d \times d}$ is the matrix for linear transformation.
The resulting $\hat{v}'$ and $V' = \{{v}'_1, {v}'_2, \dots, {v}'_{N_\text{I}}\}$ are used to replace the original image features $V$ in Eq.~(\ref{eqn:formula}) and are then fed into existing models to generate coherent reports.

In our subsequent analysis, we show that the contrastive features indeed focus on the abnormal regions and provide a better starting point for downstream models.

\begin{table*}[t]
\centering
\scriptsize
\setlength{\tabcolsep}{4pt}
\begin{tabular}{ccccccccccccccc} 
\shline
\multicolumn{1}{c|}{\multirow{2}{*}{Settings}}& \multicolumn{1}{l|}{\multirow{2}{*}{Methods}}&  \multicolumn{6}{c|}{Dataset: MIMIC-CXR \cite{Johnson2019MIMIC}} &  \multicolumn{6}{c}{Dataset: IU-X-ray \cite{Dina2016IU-Xray}}\\ \cline{3-14} 
\multicolumn{1}{c|}{} & \multicolumn{1}{c|}{} & \multicolumn{1}{c|}{B-1} &  \multicolumn{1}{c|}{B-2}         & \multicolumn{1}{c|}{B-3}         & \multicolumn{1}{c|}{B-4}         & \multicolumn{1}{c|}{M}         & \multicolumn{1}{c|}{R-L} &   \multicolumn{1}{c|}{B-1}&  \multicolumn{1}{c|}{B-2}         & \multicolumn{1}{c|}{B-3}         & \multicolumn{1}{c|}{B-4}         & \multicolumn{1}{c|}{M}         & \multicolumn{1}{c}{R-L}     \\ \hline \hline

\multicolumn{1}{c|}{\multirow{2}{*}{(a)}} &
\multicolumn{1}{l|}{NIC \cite{Vinyals2015Show}\ssymbol{2}} & \multicolumn{1}{c|}{0.290} & \multicolumn{1}{c|}{0.182} & \multicolumn{1}{c|}{0.119} & \multicolumn{1}{c|}{0.081} & \multicolumn{1}{c|}{0.112} & \multicolumn{1}{c|}{0.249} & \multicolumn{1}{c|}{0.352} & \multicolumn{1}{c|}{0.227} & \multicolumn{1}{c|}{0.154} & \multicolumn{1}{c|}{0.109} & \multicolumn{1}{c|}{0.133} & \multicolumn{1}{c }{0.313}  \\

\multicolumn{1}{c|}{} & 
\multicolumn{1}{l|}{ \ w/ Contrastive Attention} & \multicolumn{1}{c|}{\bf 0.317} & \multicolumn{1}{c|}{\bf 0.200} & \multicolumn{1}{c|}{\bf 0.127} & \multicolumn{1}{c|}{\bf 0.089} & \multicolumn{1}{c|}{\bf 0.120} & \multicolumn{1}{c|}{\bf 0.262} & \multicolumn{1}{c|}{\bf 0.368} & \multicolumn{1}{c|}{\bf 0.232} & \multicolumn{1}{c|}{\bf 0.166} & \multicolumn{1}{c|}{\bf 0.118} & \multicolumn{1}{c|}{\bf 0.144} & \multicolumn{1}{c }{\bf 0.323} \\ \hline \hline

\multicolumn{1}{c|}{\multirow{2}{*}{(b)}} &
\multicolumn{1}{l|}{Visual-Attention \cite{Xu2015Show}\ssymbol{2}} & \multicolumn{1}{c|}{\bf 0.318} & \multicolumn{1}{c|}{0.186} & \multicolumn{1}{c|}{0.122} & \multicolumn{1}{c|}{0.085} & \multicolumn{1}{c|}{0.119} & \multicolumn{1}{c|}{\bf 0.267} & \multicolumn{1}{c|}{0.371} & \multicolumn{1}{c|}{0.233} & \multicolumn{1}{c|}{0.159} & \multicolumn{1}{c|}{0.118} & \multicolumn{1}{c|}{\bf 0.147} & \multicolumn{1}{c }{\bf 0.320} \\

\multicolumn{1}{c|}{} & 
\multicolumn{1}{l|}{ \ w/ Contrastive Attention} & \multicolumn{1}{c|}{0.309} & \multicolumn{1}{c|}{\bf 0.202} & \multicolumn{1}{c|}{\bf 0.129} & \multicolumn{1}{c|}{\bf 0.093} & \multicolumn{1}{c|}{\bf 0.122} & \multicolumn{1}{c|}{0.265} & \multicolumn{1}{c|}{\bf 0.384} & \multicolumn{1}{c|}{\bf 0.245} & \multicolumn{1}{c|}{\bf 0.172} & \multicolumn{1}{c|}{\bf 0.125} & \multicolumn{1}{c|}{ 0.141} & \multicolumn{1}{c}{0.315} \\ \hline \hline

\multicolumn{1}{c|}{\multirow{2}{*}{(c)}} &
\multicolumn{1}{l|}{Spatial-Attention \cite{lu2017knowing}\ssymbol{2}}  &\multicolumn{1}{c|}{0.302} & \multicolumn{1}{c|}{0.189} & \multicolumn{1}{c|}{0.122} & \multicolumn{1}{c|}{0.082} & \multicolumn{1}{c|}{0.120} & \multicolumn{1}{c|}{0.259} & \multicolumn{1}{c|}{0.374} & \multicolumn{1}{c|}{0.235} & \multicolumn{1}{c|}{0.158} & \multicolumn{1}{c|}{\bf 0.120} & \multicolumn{1}{c|}{\bf 0.146} & \multicolumn{1}{c}{0.322} \\

\multicolumn{1}{c|}{} &
\multicolumn{1}{l|}{ \ w/ Contrastive Attention} & \multicolumn{1}{c|}{\bf 0.320} & \multicolumn{1}{c|}{\bf 0.204} & \multicolumn{1}{c|}{\bf 0.129} & \multicolumn{1}{c|}{\bf 0.091} & \multicolumn{1}{c|}{\bf 0.122} & \multicolumn{1}{c|}{\bf 0.266} & \multicolumn{1}{c|}{\bf 0.378} & \multicolumn{1}{c|}{\bf 0.236} & \multicolumn{1}{c|}{\bf 0.161} & \multicolumn{1}{c|}{0.116} & \multicolumn{1}{c|}{\bf 0.146} & \multicolumn{1}{c}{\bf 0.335} \\ \hline \hline

\multicolumn{1}{c|}{\multirow{2}{*}{(d)}} &
\multicolumn{1}{l|}{Att2in \cite{rennie2017self}\ssymbol{2}} & \multicolumn{1}{c|}{0.314} & \multicolumn{1}{c|}{0.199} & \multicolumn{1}{c|}{0.126} & \multicolumn{1}{c|}{0.087} & \multicolumn{1}{c|}{\bf 0.125} & \multicolumn{1}{c|}{0.265} & \multicolumn{1}{c|}{0.410} & \multicolumn{1}{c|}{0.257} & \multicolumn{1}{c|}{0.173} & \multicolumn{1}{c|}{0.131} & \multicolumn{1}{c|}{0.149} & \multicolumn{1}{c }{0.325}\\

\multicolumn{1}{c|}{} &
\multicolumn{1}{l|}{ \ w/ Contrastive Attention} & \multicolumn{1}{c|}{\bf 0.327} & \multicolumn{1}{c|}{\bf 0.205} & \multicolumn{1}{c|}{\bf 0.132} & \multicolumn{1}{c|}{\bf 0.095} & \multicolumn{1}{c|}{0.124} & \multicolumn{1}{c|}{\bf 0.271} & \multicolumn{1}{c|}{\bf 0.442} & \multicolumn{1}{c|}{\bf 0.281} & \multicolumn{1}{c|}{\bf 0.200} & \multicolumn{1}{c|}{\bf 0.150} & \multicolumn{1}{c|}{\bf 0.171} & \multicolumn{1}{c}{\bf 0.344} \\ \hline \hline

\multicolumn{1}{c|}{\multirow{2}{*}{(e)}} &
\multicolumn{1}{l|}{Adaptive-Attention \cite{lu2017knowing}\ssymbol{2}} &           \multicolumn{1}{c|}{0.307} & \multicolumn{1}{c|}{0.192} & \multicolumn{1}{c|}{0.124} & \multicolumn{1}{c|}{0.084} & \multicolumn{1}{c|}{0.119} & \multicolumn{1}{c|}{0.262} & \multicolumn{1}{c|}{\bf 0.433} & \multicolumn{1}{c|}{\bf 0.285} & \multicolumn{1}{c|}{0.194} & \multicolumn{1}{c|}{0.137} & \multicolumn{1}{c|}{0.166} & \multicolumn{1}{c }{\bf 0.349}\\

\multicolumn{1}{c|}{} &
\multicolumn{1}{l|}{ \ w/ Contrastive Attention} & \multicolumn{1}{c|}{\bf 0.330} & \multicolumn{1}{c|}{\bf 0.208} & \multicolumn{1}{c|}{\bf 0.134} & \multicolumn{1}{c|}{\bf 0.095} & \multicolumn{1}{c|}{\bf 0.126} & \multicolumn{1}{c|}{\bf 0.270} & \multicolumn{1}{c|}{0.425} & \multicolumn{1}{c|}{0.279} & \multicolumn{1}{c|}{\bf 0.198} & \multicolumn{1}{c|}{\bf 0.142} & \multicolumn{1}{c|}{\bf 0.167} & \multicolumn{1}{c}{0.347} \\ \hline \hline

\multicolumn{1}{c|}{\multirow{2}{*}{(f)}} &
\multicolumn{1}{l|}{Up-Down \cite{anderson2018bottom}\ssymbol{2}} & \multicolumn{1}{c|}{0.318} & \multicolumn{1}{c|}{0.203} & \multicolumn{1}{c|}{0.128} & \multicolumn{1}{c|}{0.089} & \multicolumn{1}{c|}{0.123} & \multicolumn{1}{c|}{0.266} & \multicolumn{1}{c|}{\bf 0.389} & \multicolumn{1}{c|}{\bf 0.251} & \multicolumn{1}{c|}{\bf 0.170} & \multicolumn{1}{c|}{0.126} & \multicolumn{1}{c|}{\bf 0.154} & \multicolumn{1}{c }{0.317}\\

\multicolumn{1}{c|}{} &
\multicolumn{1}{l|}{ \ w/ Contrastive Attention} & \multicolumn{1}{c|}{\bf 0.336} & \multicolumn{1}{c|}{\bf 0.209} & \multicolumn{1}{c|}{\bf 0.134} & \multicolumn{1}{c|}{\bf 0.097} & \multicolumn{1}{c|}{\bf 0.128} & \multicolumn{1}{c|}{\bf 0.273} & \multicolumn{1}{c|}{0.378} & \multicolumn{1}{c|}{0.246} & \multicolumn{1}{c|}{0.169} & \multicolumn{1}{c|}{\bf 0.129} & \multicolumn{1}{c|}{0.152} & \multicolumn{1}{c}{\bf 0.330} \\ \hline \hline

\multicolumn{1}{c|}{\multirow{2}{*}{(g)}} &
\multicolumn{1}{l|}{HLSTM \cite{Krause2017Hierarchical}\ssymbol{2}} & \multicolumn{1}{c|}{0.321} & \multicolumn{1}{c|}{0.203} & \multicolumn{1}{c|}{0.129} & \multicolumn{1}{c|}{0.092} & \multicolumn{1}{c|}{0.125} & \multicolumn{1}{c|}{0.270} & \multicolumn{1}{c|}{0.435} & \multicolumn{1}{c|}{0.280} & \multicolumn{1}{c|}{0.187} & \multicolumn{1}{c|}{0.131} & \multicolumn{1}{c|}{0.173} & \multicolumn{1}{c}{0.346} \\

\multicolumn{1}{c|}{} &
\multicolumn{1}{l|}{ \ w/ Contrastive Attention} & \multicolumn{1}{c|}{\bf 0.352} & \multicolumn{1}{c|}{\bf 0.216} & \multicolumn{1}{c|}{\bf 0.145} & \multicolumn{1}{c|}{\bf 0.105} & \multicolumn{1}{c|}{\bf 0.139} & \multicolumn{1}{c|}{\bf 0.276} & \multicolumn{1}{c|}{\bf 0.453} & \multicolumn{1}{c|}{\bf 0.290} & \multicolumn{1}{c|}{\bf 0.203} & \multicolumn{1}{c|}{\bf 0.153} & \multicolumn{1}{c|}{\bf 0.178} & \multicolumn{1}{c}{\bf 0.361} \\
\hline \hline

\multicolumn{1}{c|}{\multirow{2}{*}{(h)}} &                   
\multicolumn{1}{l|}{HLSTM+att+Dual  \cite{Harzig2019Addressing}\ssymbol{2}}     & \multicolumn{1}{c|}{\bf 0.328} & \multicolumn{1}{c|}{\bf 0.204} & \multicolumn{1}{c|}{0.127} & \multicolumn{1}{c|}{0.090} & \multicolumn{1}{c|}{0.122} & \multicolumn{1}{c|}{0.267} & \multicolumn{1}{c|}{0.447} & \multicolumn{1}{c|}{0.289} & \multicolumn{1}{c|}{0.192} & \multicolumn{1}{c|}{0.144} & \multicolumn{1}{c|}{0.175} & \multicolumn{1}{c }{0.358}\\

\multicolumn{1}{c|}{} & 
\multicolumn{1}{l|}{ \ w/ Contrastive Attention} & \multicolumn{1}{c|}{0.323} & \multicolumn{1}{c|}{0.202} & \multicolumn{1}{c|}{\bf 0.130} & \multicolumn{1}{c|}{\bf 0.102} & \multicolumn{1}{c|}{\bf 0.138} & \multicolumn{1}{c|}{\bf 0.277} & \multicolumn{1}{c|}{\bf 0.464} & \multicolumn{1}{c|}{\bf 0.292} & \multicolumn{1}{c|}{\bf 0.205} & \multicolumn{1}{c|}{\bf 0.149} & \multicolumn{1}{c|}{\bf 0.176} & \multicolumn{1}{c}{\bf 0.364}\\
\hline \hline

\multicolumn{1}{c|}{\multirow{2}{*}{(i)}} &
\multicolumn{1}{l|}{Co-Attention \cite{Jing2018Automatic}\ssymbol{2}}& \multicolumn{1}{c|}{0.329} & \multicolumn{1}{c|}{0.206} & \multicolumn{1}{c|}{0.133} & \multicolumn{1}{c|}{0.095} & \multicolumn{1}{c|}{0.129} & \multicolumn{1}{c|}{\bf 0.273} & \multicolumn{1}{c|}{0.463} & \multicolumn{1}{c|}{0.293} & \multicolumn{1}{c|}{0.207} & \multicolumn{1}{c|}{0.155} & \multicolumn{1}{c|}{0.178} & \multicolumn{1}{c }{0.365} \\

\multicolumn{1}{c|}{} & 
\multicolumn{1}{l|}{ \ w/ Contrastive Attention} & \multicolumn{1}{c|}{\bf 0.351} & \multicolumn{1}{c|}{\bf 0.213} & \multicolumn{1}{c|}{\bf 0.148} & \multicolumn{1}{c|}{\bf 0.106} & \multicolumn{1}{c|}{\bf 0.147} & \multicolumn{1}{c|}{0.270} & \multicolumn{1}{c|}{\bf 0.486} & \multicolumn{1}{c|}{\bf 0.311} & \multicolumn{1}{c|}{\bf\color{red}  0.223} & \multicolumn{1}{c|}{\bf\color{red} 0.178} & \multicolumn{1}{c|}{\bf 0.187} & \multicolumn{1}{c}{\bf 0.372}\\
\hline \hline

\multicolumn{1}{c|}{\multirow{2}{*}{(j)}} &
\multicolumn{1}{l|}{Multi-Attention \cite{Huang2019Multi_Attention}\ssymbol{2}}                  &\multicolumn{1}{c|}{0.337} & \multicolumn{1}{c|}{0.211} & \multicolumn{1}{c|}{0.136} & \multicolumn{1}{c|}{0.097} & \multicolumn{1}{c|}{0.130} & \multicolumn{1}{c|}{0.274} & \multicolumn{1}{c|}{0.468} & \multicolumn{1}{c|}{0.299} & \multicolumn{1}{c|}{0.211} & \multicolumn{1}{c|}{0.155} & \multicolumn{1}{c|}{0.180} & \multicolumn{1}{c}{0.366}\\

\multicolumn{1}{c|}{} &
\multicolumn{1}{l|}{ \ w/ Contrastive Attention} & \multicolumn{1}{c|}{\bf\color{red} 0.350} & \multicolumn{1}{c|}{\bf\color{red} 0.219} & \multicolumn{1}{c|}{\bf\color{red} 0.152} & \multicolumn{1}{c|}{\bf\color{red} 0.109} & \multicolumn{1}{c|}{\bf\color{red} 0.151} & \multicolumn{1}{c|}{\bf\color{red} 0.283} & \multicolumn{1}{c|}{\bf\color{red} 0.492} & \multicolumn{1}{c|}{\bf\color{red} 0.314} & \multicolumn{1}{c|}{\bf 0.222} & \multicolumn{1}{c|}{\bf 0.169} & \multicolumn{1}{c|}{\bf\color{red} 0.193} & \multicolumn{1}{c}{\bf\color{red} 0.381}\\
\hline

\shline
\end{tabular}
\caption{Performance of automatic evaluations on the test set of the MIMIC-CXR dataset and the IU-X-ray dataset. \ssymbol{2} denotes our own implementation. B-n, M and R-L are short for BLEU-n, METEOR and ROUGE-L, respectively.  Higher is better in all columns. In this paper, the {\color{red} Red} colored numbers denote the best results across all approaches in Table. As we can see, most baseline models enjoy a comfortable improvement with our approach.}
\label{tab:automatic}
\end{table*}

\begin{table*}[t]
\centering
\scriptsize
\begin{tabular}{cccccccccccccc} 
\shline
\multicolumn{1}{l|}{\multirow{2}{*}{Methods}}&  \multicolumn{6}{c|}{Dataset: MIMIC-CXR \cite{Johnson2019MIMIC}} &  \multicolumn{6}{c}{Dataset: IU-X-ray \cite{Dina2016IU-Xray}}\\ \cline{2-13} 
\multicolumn{1}{c|}{} & \multicolumn{1}{c|}{B-1} &  \multicolumn{1}{c|}{B-2}         & \multicolumn{1}{c|}{B-3}         & \multicolumn{1}{c|}{B-4}         & \multicolumn{1}{c|}{M}         & \multicolumn{1}{c|}{R-L} &   \multicolumn{1}{c|}{B-1}&  \multicolumn{1}{c|}{B-2}         & \multicolumn{1}{c|}{B-3}         & \multicolumn{1}{c|}{B-4}         & \multicolumn{1}{c|}{M}         & \multicolumn{1}{c}{R-L}     \\ \hline \hline

\multicolumn{1}{l|}{HRGR-Agent \cite{Li2018Hybrid}} & \multicolumn{1}{c|}{-} & \multicolumn{1}{c|}{-} & \multicolumn{1}{c|}{-} & \multicolumn{1}{c|}{-} & \multicolumn{1}{c|}{-} & \multicolumn{1}{c|}{-} & \multicolumn{1}{c|}{0.438} & \multicolumn{1}{c|}{0.298} & \multicolumn{1}{c|}{0.208} & \multicolumn{1}{c|}{0.151} & \multicolumn{1}{c|}{-} & \multicolumn{1}{c}{0.322}  \\

\multicolumn{1}{l|}{CMAS-RL \cite{Jing2019Show}} & \multicolumn{1}{c|}{-} & \multicolumn{1}{c|}{-} & \multicolumn{1}{c|}{-} & \multicolumn{1}{c|}{-} & \multicolumn{1}{c|}{-} & \multicolumn{1}{c|}{-} & \multicolumn{1}{c|}{0.464} & \multicolumn{1}{c|}{0.301} & \multicolumn{1}{c|}{0.210} & \multicolumn{1}{c|}{0.154} & \multicolumn{1}{c|}{-} & \multicolumn{1}{c}{0.362}  \\

\multicolumn{1}{l|}{SentSAT + KG \cite{Zhang2020When}} & \multicolumn{1}{c|}{-} & \multicolumn{1}{c|}{-} & \multicolumn{1}{c|}{-} & \multicolumn{1}{c|}{-} & \multicolumn{1}{c|}{-} & \multicolumn{1}{c|}{-} & \multicolumn{1}{c|}{0.441} & \multicolumn{1}{c|}{0.291} & \multicolumn{1}{c|}{0.203} & \multicolumn{1}{c|}{0.147} & \multicolumn{1}{c|}{-} & \multicolumn{1}{c}{0.367}  \\

\multicolumn{1}{l|}{Transformer \cite{Chen2020Generating}} & \multicolumn{1}{c|}{0.314} & \multicolumn{1}{c|}{0.192} & \multicolumn{1}{c|}{0.127} & \multicolumn{1}{c|}{0.090} & \multicolumn{1}{c|}{0.125} & \multicolumn{1}{c|}{0.265} & \multicolumn{1}{c|}{0.396} &      \multicolumn{1}{c|}{0.254} & \multicolumn{1}{c|}{0.179} & \multicolumn{1}{c|}{0.135} & \multicolumn{1}{c|}{0.164} & \multicolumn{1}{c}{0.342}  \\     

\multicolumn{1}{l|}{R2Gen \cite{Chen2020Generating}} & \multicolumn{1}{c|}{\color{red} 0.353} & \multicolumn{1}{c|}{0.218} & \multicolumn{1}{c|}{0.145} & \multicolumn{1}{c|}{0.103} & \multicolumn{1}{c|}{0.142} & \multicolumn{1}{c|}{0.277} & \multicolumn{1}{c|}{0.470} & \multicolumn{1}{c|}{0.304} & \multicolumn{1}{c|}{0.219} & \multicolumn{1}{c|}{0.165} & \multicolumn{1}{c|}{0.187} & \multicolumn{1}{c}{0.371} \\
\hline \hline

\multicolumn{1}{l|}{Contrastive Attention (Ours)} & \multicolumn{1}{c|}{0.350} & \multicolumn{1}{c|}{\color{red} 0.219} & \multicolumn{1}{c|}{\color{red} 0.152} & \multicolumn{1}{c|}{\color{red} 0.109} & \multicolumn{1}{c|}{\color{red} 0.151} & \multicolumn{1}{c|}{\color{red} 0.283} & \multicolumn{1}{c|}{\color{red} 0.492} & \multicolumn{1}{c|}{\color{red} 0.314} & \multicolumn{1}{c|}{\color{red} 0.222} & \multicolumn{1}{c|}{\color{red} 0.169} & \multicolumn{1}{c|}{\color{red} 0.193} & \multicolumn{1}{c}{\color{red} 0.381}\\
\hline

\shline
\end{tabular}
\caption{Comparison with existing state-of-the-art methods on the test set of the MIMIC-CXR dataset and the IU-X-ray dataset. As we can see, we achieve the state-of-the-art performance on major metrics on the two datasets.}
\label{tab:SOTA}
\end{table*}

\section{Experiments}
\label{sec:experiments}

\subsection{Datasets, Baselines and Settings}

\smallskip\noindent\textbf{Datasets} \
We use the widely-used benchmark IU-X-ray
\cite{Dina2016IU-Xray} dataset and the recently released large-scale MIMIC-CXR
\cite{Johnson2019MIMIC} datasets for evaluation.

    $\bullet$ \textbf{IU-X-ray}\footnote{\url{https://openi.nlm.nih.gov/}} contains 3,955 radiology reports and 7,470 X-rays images. We use the widely-used splits in \citet{Chen2020Generating,Jing2019Show,Li2019Knowledge,Li2018Hybrid,fenglin2021PPKED} for evaluation. There are 70\%, 10\% and 20\% instances in training set, validation set and test set, respectively. 
    
    $\bullet$ \textbf{MIMIC-CXR}\footnote{\url{https://physionet.org/content/mimic-cxr/2.0.0/}} is the recently released largest dataset to date and consists of 377,110 chest X-ray images and 227,835 reports from 64,588 patients. Following \citet{Chen2020Generating,fenglin2021PPKED}, we use the official splits to report our results. Thus, there are 368,960 in the training set, 2,991 in the validation set and 5,159 in the test set. We convert all tokens of reports to lower-cases and remove the tokens whose frequency of occurrence in the training set is less than 10, resulting in 4k words.

\smallskip\noindent\textbf{Baselines} \
We experiment with two lines of baselines that are originally designed for image captioning and chest X-ray report generation.

    $\bullet$ \textbf{Image Captioning Baselines} \ We experiment on six representative models: NIC \cite{Vinyals2015Show}, Visual-Attention \cite{Xu2015Show}, Spatial-Attention \cite{lu2017knowing}, Att2in \cite{rennie2017self}, Adaptive-Attention \cite{lu2017knowing} and Up-Down \cite{anderson2018bottom}.
    
    $\bullet$ \textbf{Chest X-ray Report Generation Baselines} \ We conduct the experiment on four baselines consisting of HLSTM \cite{Krause2017Hierarchical,Jing2018Automatic}, HLSTM+att+Dual \cite{Harzig2019Addressing}, Co-Attention \cite{Jing2018Automatic} and Multi-Attention \cite{Huang2019Multi_Attention}. 

\smallskip\noindent\textbf{Settings} \
For our Contrastive Attention model, the model size $d$ is set to 512.
Based on the average performance on the validation set, the $n$ in the Aggregate Attention is set to 6.
For the normality pool, we randomly extract 1,000 normal images, i.e., $N_\text{P}=1,000$, for both two datasets from their training set.
To re-implement the baselines, following \citet{Jing2019Show,Li2019Knowledge,Li2018Hybrid,fenglin2021PPKED}, we adopt the ResNet-50 \cite{he2016deep} pre-trained on ImageNet \cite{Deng2009ImageNet} and fine-tuned on CheXpert dataset \cite{Irvin2019CheXpert} to extract the patch visual features with the dimension of each feature is 2,048, which will be projected to 512.
Besides, we utilize paired images of a patient as the input for IU-X-ray and utilize single image as the input for MIMIC-CXR to ensure consistency with the experiment settings of previous works \cite{Chen2020Generating}.
For all baselines, since our focus is to provide explicit abnormal region features, which tends to improve existing baselines, we keep the inner structure of the baselines untouched and preserve the original parameter setting and training strategy.

\begin{table*}[t]
\centering
\scriptsize
\setlength{\tabcolsep}{3pt}
\begin{tabular}{llccccccc} 
\shline
\multicolumn{1}{l|}{\multirow{2}{*}{Metrics}}& \multicolumn{1}{l|}{\multirow{2}{*}{vs. Models}} & \multicolumn{3}{c|}{Dataset: MIMIC-CXR \cite{Johnson2019MIMIC}} &  \multicolumn{3}{c}{Dataset: IU-X-ray \cite{Dina2016IU-Xray}}\\ 
\cline{3-8} 
\multicolumn{1}{c|}{} &\multicolumn{1}{c|}{} &\multicolumn{1}{c|}{Baseline wins}&  \multicolumn{1}{c|}{Tie}         & \multicolumn{1}{c|}{`w/ CA' wins}         & \multicolumn{1}{c|}{Baseline wins}         & \multicolumn{1}{c|}{Tie}         & \multicolumn{1}{c}{`w/ CA' wins}     \\ \hline \hline

\multicolumn{1}{l|}{\multirow{2}{*}{Fluency}} & \multicolumn{1}{l|}{HLSTM \cite{Krause2017Hierarchical}\ssymbol{2}} & \multicolumn{1}{c|}{21} & \multicolumn{1}{c|}{45} & \multicolumn{1}{c|}{\bf 34} & \multicolumn{1}{c|}{18} & \multicolumn{1}{c|}{43} & \multicolumn{1}{c}{\bf 39} \\
\multicolumn{1}{l|}{} & \multicolumn{1}{l|}{Multi-Attention \cite{Huang2019Multi_Attention}\ssymbol{2}} & \multicolumn{1}{c|}{30} & \multicolumn{1}{c|}{34} & \multicolumn{1}{c|}{\bf 36} & \multicolumn{1}{c|}{27} & \multicolumn{1}{c|}{41} & \multicolumn{1}{c}{\bf 32} \\
\hline  \hline

\multicolumn{1}{l|}{\multirow{2}{*}{Comprehensiveness}} & \multicolumn{1}{l|}{HLSTM \cite{Krause2017Hierarchical}\ssymbol{2}} & \multicolumn{1}{c|}{13} & \multicolumn{1}{c|}{33} & \multicolumn{1}{c|}{\bf 54} & \multicolumn{1}{c|}{10} & \multicolumn{1}{c|}{19} & \multicolumn{1}{c}{\bf 71} \\
\multicolumn{1}{l|}{} & \multicolumn{1}{l|}{Multi-Attention \cite{Huang2019Multi_Attention}\ssymbol{2}} & \multicolumn{1}{c|}{26} & \multicolumn{1}{c|}{35} & \multicolumn{1}{c|}{\bf 39} & \multicolumn{1}{c|}{22} & \multicolumn{1}{c|}{31} & \multicolumn{1}{c}{\bf 47} \\
\hline  \hline

\multicolumn{1}{l|}{\multirow{2}{*}{Faithfulness}} & \multicolumn{1}{l|}{HLSTM \cite{Krause2017Hierarchical}\ssymbol{2}} & \multicolumn{1}{c|}{20} & \multicolumn{1}{c|}{28} & \multicolumn{1}{c|}{\bf 52} & \multicolumn{1}{c|}{13} & \multicolumn{1}{c|}{25} & \multicolumn{1}{c}{\bf 62} \\
\multicolumn{1}{l|}{} & \multicolumn{1}{l|}{Multi-Attention \cite{Huang2019Multi_Attention}\ssymbol{2}} & \multicolumn{1}{c|}{31} & \multicolumn{1}{c|}{32} & \multicolumn{1}{c|}{\bf 37} & \multicolumn{1}{c|}{24} & \multicolumn{1}{c|}{30} & \multicolumn{1}{c}{\bf 46} \\
\hline  

\shline
\end{tabular}
\caption{Results of human evaluation on MIMIC-CXR and IU-X-ray datasets for comparing our method with baselines in terms of the 
\textbf{fluency} of generated reports, the \textbf{comprehensiveness} of the generated true abnormalities and the \textbf{faithfulness} to the ground truth reports. All values are reported in percentage (\%).
\label{tab:human_evaluation}}
\end{table*}
\subsection{Main Evaluation}

\smallskip\noindent\textbf{Metrics} \ 
We first perform the automatic evaluation to conduct a fair comparison. To measure performance, we adopt the widely-used evaluation toolkit \cite{chen2015microsoft} to calculate the standard metrics: BLEU \cite{papineni2002bleu}, METEOR \cite{Banerjee2005METEOR} and ROUGE-L \cite{lin2004rouge}.
Specifically, BLEU and METEOR are originally designed for machine translation evaluation, while ROUGE is originally proposed for automatic evaluation of the extracted text summarization.

\smallskip\noindent\textbf{Results} \
The results on the test set of MIMIC-CXR and IU-X-ray datasets are reported in Table~\ref{tab:automatic}.
As we can see, our Contrastive Attention can successfully boost baselines with improvement up to 14\% and 17\% for MIMIC-CXR and IU-X-ray in terms of BLEU-4 score, respectively,
where the Setting (g) achieves the greatest improvements.
The results prove the effectiveness and generalization capabilities of our approach to a wide range of models.

Moreover, in Table~\ref{tab:SOTA}, we choose five competitive models including the current state-of-the-art models, i.e., SentSAT + KG \cite{Zhang2020When} and R2Gen \cite{Chen2020Generating}, for comparison.
For these competitive models, we directly report the results from the original paper.
Table~\ref{tab:SOTA} shows that when our approach is applied to the Multi-Attention, we outperform these existing state-of-the-art models on major metrics on the IU-X-ray and MIMIC-CXR datasets, which further proves the effectiveness of our Contrastive Attention.

\subsection{Clinical Efficacy}
\label{sec:automatic_CE}
\smallskip\noindent\textbf{Metrics} \ 
The metrics used in the Table~\ref{tab:automatic} measure the match between the generated reports and ground truth reports, but are not specialized for the abnormalities in the reports.
Therefore, to measure the accuracy of descriptions for clinical abnormalities, we follow \citet{Chen2020Generating} to adopt the CheXpert labeler \cite{Irvin2019CheXpert}, which will label the given reports in 14 different categories related to thoracic diseases and support devices, to further report the clinical efficacy metrics.
As a result, we can calculate the clinical efficacy scores by comparing the generated reports with ground truth reports in 14 different categories, producing the Precision, Recall and F1 scores.

\smallskip\noindent\textbf{Results} \
The results are shown in Table~\ref{tab:automatic_CE}. As we can see, our approach can boost the performance of baselines under all clinical efficacy metrics.
The results prove our arguments and the effectiveness of our approach in boosting the baselines to correctly capture and depict the abnormalities.
Specifically, our Multi-Attention w/ CA outperforms the R2Gen \cite{Chen2020Generating} with relatively 6\%, 9\% and 10\% margins in terms of Precision, Recall and F1 scores, respectively
The superior clinical efficacy scores, which measure the accuracy of descriptions for clinical abnormalities, demonstrate that our approach can help existing models produce higher quality descriptions for clinical abnormalities.

\begin{table}[t]
    \centering
    \scriptsize
    \begin{tabular}{@{}l c c c @{}}
    \toprule
    
    \multirow{2}{*}[-3pt]{Methods} & \multicolumn{3}{c}{Clinical Efficacy  Metrics}  \\ \cmidrule(l){2-4} 
    & Precision & Recall & F1   \\
    \midrule
    NIC \cite{Vinyals2015Show}   &  0.249 & 0.203&  0.204 \\
    AdaAtt \cite{lu2017knowing} &  0.268 & 0.186&  0.181\\
    Att2in \cite{rennie2017self}  & 0.322 & 0.239&  0.249 \\ 
    Up-Down \cite{anderson2018bottom} & 0.320 & 0.231&  0.238\\
    Transformer \cite{Chen2020Generating}    &0.331 &0.224 &0.228 \\
    R2Gen \cite{Chen2020Generating}   &0.333 &0.273 &0.276
    \\ \midrule
    
    HLSTM \cite{Krause2017Hierarchical}\ssymbol{2} & 0.307 & 0.225 & 0.228\\
     \ w/ Contrastive Attention & \bf 0.340 & \bf 0.269  & \bf 0.274  \\ \midrule
     
    Multi-Attention \cite{Huang2019Multi_Attention}\ssymbol{2} &0.336 &0.257 &0.265\\
     \ w/ Contrastive Attention & \color{red}  \bf 0.352 &\color{red}  \bf 0.298 &\color{red}  \bf 0.303
    \\ 
    \bottomrule
    \end{tabular}
    \caption{Performance of automatic evaluations in terms of clinical efficacy metrics, which measure the accuracy of descriptions for clinical abnormalities, on the MIMIC-CXR dataset. Higher is better in all columns.}
    \label{tab:automatic_CE}
\end{table}

\begin{table*}[t]
\centering
\scriptsize
\setlength{\tabcolsep}{4pt}   
\begin{tabular}{ccccccccccccccccc}
\shline
\multicolumn{1}{c|}{\multirow{3}{*}{Settings}} & \multicolumn{1}{l|}{\multirow{3}{*}{Methods}} & \multicolumn{1}{c|}{\multirow{3}{*}{Attention Function}} & \multicolumn{1}{l|}{\multirow{3}{*}{$n$}} &  \multicolumn{12}{c}{Dataset: IU-X-ray \cite{Dina2016IU-Xray}}
 \\  \cline{5-16}
 
\multicolumn{1}{c|}{} & \multicolumn{1}{c|}{} & \multicolumn{1}{c|}{} & \multicolumn{1}{c|}{} &  \multicolumn{6}{c|}{Baseline: HLSTM \cite{Krause2017Hierarchical}} &  \multicolumn{6}{c}{Baseline: Multi-Attention \cite{Huang2019Multi_Attention}} \\ \cline{5-16} 
\multicolumn{1}{c|}{} & \multicolumn{1}{c|}{} & \multicolumn{1}{c|}{} & \multicolumn{1}{c|}{}            & \multicolumn{1}{c|}{B-1}         &  \multicolumn{1}{c|}{B-2}         & \multicolumn{1}{c|}{B-3}         & \multicolumn{1}{c|}{B-4}         & \multicolumn{1}{c|}{M}           & \multicolumn{1}{c|}{R-L}         &   \multicolumn{1}{c|}{B-1}         &  \multicolumn{1}{c|}{B-2}         & \multicolumn{1}{c|}{B-3}         & \multicolumn{1}{c|}{B-4}         & \multicolumn{1}{c|}{M}           & \multicolumn{1}{c}{R-L}     \\ \hline \hline

\multicolumn{1}{c|}{(a)} &\multicolumn{1}{l|}{Baseline} & \multicolumn{1}{c|}{-} 
& \multicolumn{1}{c|}{-}     & \multicolumn{1}{c|}{0.435}   & \multicolumn{1}{c|}{0.280}   & \multicolumn{1}{c|}{0.187}   & \multicolumn{1}{c|}{0.131}   & \multicolumn{1}{c|}{0.173}   & \multicolumn{1}{c|}{0.346}   & \multicolumn{1}{c|}{0.468}   & \multicolumn{1}{c|}{0.299}   & \multicolumn{1}{c|}{0.211}   & \multicolumn{1}{c|}{0.155}   & \multicolumn{1}{c|}{0.180}   &  \multicolumn{1}{c}{0.366} \\ \hline \hline

\multicolumn{1}{c|}{(b)} &\multicolumn{1}{l|}{\ w/ DA} & \multicolumn{1}{c|}{\multirow{2}{*}{dot-product attention}} & \multicolumn{1}{c|}{-} & \multicolumn{1}{c|}{0.447}   & \multicolumn{1}{c|}{0.284}   & \multicolumn{1}{c|}{0.195}   & \multicolumn{1}{c|}{0.138}   & \multicolumn{1}{c|}{0.176}   & \multicolumn{1}{c|}{0.357}   & \multicolumn{1}{c|}{0.479}   & \multicolumn{1}{c|}{0.306}   & \multicolumn{1}{c|}{0.215}   & \multicolumn{1}{c|}{0.162}   & \multicolumn{1}{c|}{0.184}  & \multicolumn{1}{c}{0.373}\\
\multicolumn{1}{c|}{(c)} &\multicolumn{1}{l|}{\  w/ DA + AA} & \multicolumn{1}{c|}{} & \multicolumn{1}{c|}{1}  & \multicolumn{1}{c|}{0.451} & \multicolumn{1}{c|}{0.287} & \multicolumn{1}{c|}{0.198} & \multicolumn{1}{c|}{0.144} & \multicolumn{1}{c|}{0.176} & \multicolumn{1}{c|}{0.359} & \multicolumn{1}{c|}{0.486} & \multicolumn{1}{c|}{0.310} & \multicolumn{1}{c|}{0.218} & \multicolumn{1}{c|}{0.164} & \multicolumn{1}{c|}{0.187} & \multicolumn{1}{c}{0.375}\\ \hline \hline

\multicolumn{1}{c|}{(f)} & \multicolumn{1}{l|}{\  w/ DA + AA} & \multicolumn{1}{c|}{\multirow{5}{*}{dot-product attention}} & \multicolumn{1}{c|}{2} & \multicolumn{1}{c|}{0.451} & \multicolumn{1}{c|}{0.285} & \multicolumn{1}{c|}{0.199} & \multicolumn{1}{c|}{0.146} & \multicolumn{1}{c|}{0.175} & \multicolumn{1}{c|}{0.359} & \multicolumn{1}{c|}{0.488} & \multicolumn{1}{c|}{0.312} & \multicolumn{1}{c|}{0.219} & \multicolumn{1}{c|}{0.165} & \multicolumn{1}{c|}{0.189} & \multicolumn{1}{c}{0.377} \\
\multicolumn{1}{c|}{(g)} & \multicolumn{1}{l|}{\  w/ DA + AA} & \multicolumn{1}{c|}{} & \multicolumn{1}{c|}{4} & \multicolumn{1}{c|}{\color{red} 0.453} & \multicolumn{1}{c|}{0.287} & \multicolumn{1}{c|}{0.202} & \multicolumn{1}{c|}{0.149} & \multicolumn{1}{c|}{0.177} & \multicolumn{1}{c|}{0.361} & \multicolumn{1}{c|}{\color{red} 0.493} & \multicolumn{1}{c|}{\color{red} 0.314} & \multicolumn{1}{c|}{0.220} & \multicolumn{1}{c|}{0.165} & \multicolumn{1}{c|}{0.191} & \multicolumn{1}{c}{0.378}\\
\multicolumn{1}{c|}{(h)} & 
\multicolumn{1}{l|}{\  w/ DA + AA} & \multicolumn{1}{c|}{}       & \multicolumn{1}{c|}{6}      & \multicolumn{1}{c|}{\color{red} 0.453} & \multicolumn{1}{c|}{\color{red} 0.290} & \multicolumn{1}{c|}{\color{red} 0.203} & \multicolumn{1}{c|}{\color{red} 0.153} & \multicolumn{1}{c|}{0.178} & \multicolumn{1}{c|}{0.361} & \multicolumn{1}{c|}{0.492} & \multicolumn{1}{c|}{\color{red} 0.314} & \multicolumn{1}{c|}{\color{red} 0.222} & \multicolumn{1}{c|}{\color{red} 0.169} & \multicolumn{1}{c|}{0.193} & \multicolumn{1}{c}{\color{red} 0.381} \\
\multicolumn{1}{c|}{(i)} & \multicolumn{1}{l|}{\  w/ DA + AA} & \multicolumn{1}{c|}{} & \multicolumn{1}{c|}{8} & \multicolumn{1}{c|}{0.451} & \multicolumn{1}{c|}{0.288} & \multicolumn{1}{c|}{0.200} & \multicolumn{1}{c|}{0.150} & \multicolumn{1}{c|}{\color{red} 0.180} & \multicolumn{1}{c|}{\color{red} 0.362} & \multicolumn{1}{c|}{0.488} & \multicolumn{1}{c|}{0.311} & \multicolumn{1}{c|}{0.221} & \multicolumn{1}{c|}{0.164} & \multicolumn{1}{c|}{\color{red} 0.194} & \multicolumn{1}{c}{0.379}\\
\multicolumn{1}{c|}{(j)} & \multicolumn{1}{l|}{\  w/ DA + AA} & \multicolumn{1}{c|}{}& \multicolumn{1}{c|}{10} & \multicolumn{1}{c|}{0.448} & \multicolumn{1}{c|}{0.284} & \multicolumn{1}{c|}{0.197} & \multicolumn{1}{c|}{0.142} & \multicolumn{1}{c|}{0.175} & \multicolumn{1}{c|}{0.355} & \multicolumn{1}{c|}{0.490} & \multicolumn{1}{c|}{0.310} & \multicolumn{1}{c|}{0.219} & \multicolumn{1}{c|}{0.162} & \multicolumn{1}{c|}{0.188} & \multicolumn{1}{c}{0.375} \\
\hline 
\shline
\end{tabular}
\caption{Quantitative analysis of our Contrastive Attention, which includes the Differentiate Attention (DA) and Aggregate Attention (AA). We conduct the analysis on a widely-used baseline model HLSTM \cite{Krause2017Hierarchical} and a competitive baseline model Multi-Attention \cite{Huang2019Multi_Attention}.}
\label{tab:ablation}
\end{table*}

\subsection{Human Evaluation} 
\label{sec:human_evaluation}
\smallskip\noindent\textbf{Metrics} \  
For medical-related task,
it is important to know (1) on what fraction of images with abnormalities did the system not mention the abnormality and (2) on what fraction of images the system described abnormality that does not exist according to doctors.
To this end, we randomly select 200 samples from the IU-X-ray and MIMIC-CXR test sets, which are 100 samples from each dataset. 
Specifically, it is important to generate accurate reports (\textit{faithfulness}) with comprehensive true abnormalities (\textit{comprehensiveness}) and it is unacceptable to generate repeated sentences (\textit{fluency}).
Therefore, we invite several professional clinicians to compare our approach and baselines independently and evaluate the perceptual quality, including the \textit{fluency} of generated reports, the \textit{comprehensiveness} of generated true abnormalities and the \textit{faithfulness} to the ground truth reports.
The clinicians are unaware of which model generates these reports.

\smallskip\noindent\textbf{Results} \
To conduct the human evaluation, we select 
a representative chest X-ray report generation baseline: HLSTM \cite{Krause2017Hierarchical} and a competitive chest X-ray report generation baseline: Multi-Attention \cite{Huang2019Multi_Attention}.
The results in Table~\ref{tab:human_evaluation} show that our approach enjoys the obvious advantage in terms of the three aspects, i.e., \textit{fluency}, \textit{comprehensiveness} and \textit{faithfulness}, meaning that the reports generated by the ``Baseline w/ Contrastive Attention'' are of higher clinical quality, which also proves the advantage of our approach in clinical practice.
Especially, by using our proposed Contrastive Attention, the winning chances of models increased by maximum of $54-13=41$ points and $71-10=61$ points in terms of the \textit{comprehensiveness} metric on the MIMIC-CXR and IU-X-ray datasets, respectively. It demonstrates the effectiveness of our approach in helping existing baselines generate more accurate abnormality descriptions, and thus improve the usefulness of models in better assisting radiologists in clinical decision-makings and reducing their workload.

\smallskip\noindent\textbf{Overall} \ From the results of automatic and human evaluations, we can see that our proposed Contrastive Attention can provide a solid basis for describing chest X-ray images, especially for the abnormalities. 
As a result, our approach can successfully boost baselines and achieves new state-of-the-art results on the MIMIC-CXR and IU-X-ray datasets, which verifies the effectiveness of the proposed approach and indicates that our approach is less prone to the variations of model structures, hyper-parameters (e.g., learning rate and batch size), and learning paradigms.

\section{Analysis}
\label{sec:analysis}
We conduct analysis on the benchmark IU-X-ray dataset to better understand our proposed approach.

\subsection{Quantitative Analysis}
We conduct the quantitative analysis on two representative models, i.e., HLSTM and Multi-Attention, to evaluate the contribution of each component. 

\smallskip\noindent\textbf{Effect of Contrastive Attention} \
Our Contrastive Attention consists of the Differentiate Attention (DA) and Aggregate Attention (AA).
As shown in Table~\ref{tab:ablation}(b), the DA can promote the performance over all metrics.
Especially for the HLSTM \cite{Krause2017Hierarchical}, which does not incorporate the attention mechanism to allow more efficient use of the image features. We can see that an up to 5\% gain in BLEU-4 score makes the ``HLSTM w/ DA'' an equally competitive model as the ``HLSTM+att+Dual'' model in Table~\ref{tab:automatic}(h).
This indicates that the contrastive information extracted by the DA contains sufficient accurate abnormal information, which is vital in improving the performance of chest X-ray report generation.
In other words, our approach can ease the design of the neural models for the task.

For the AA, it devotes to identifying the closest normal images and filtering out the noisy images, which can improve the contrasting process in the DA.
As expected, Table~\ref{tab:ablation}(c) shows that the AA can consistently boost the performance of baselines under all metrics, which further demonstrates the effectiveness of our approach.

\begin{figure*}[t]

\centering
\includegraphics[width=0.995\linewidth]{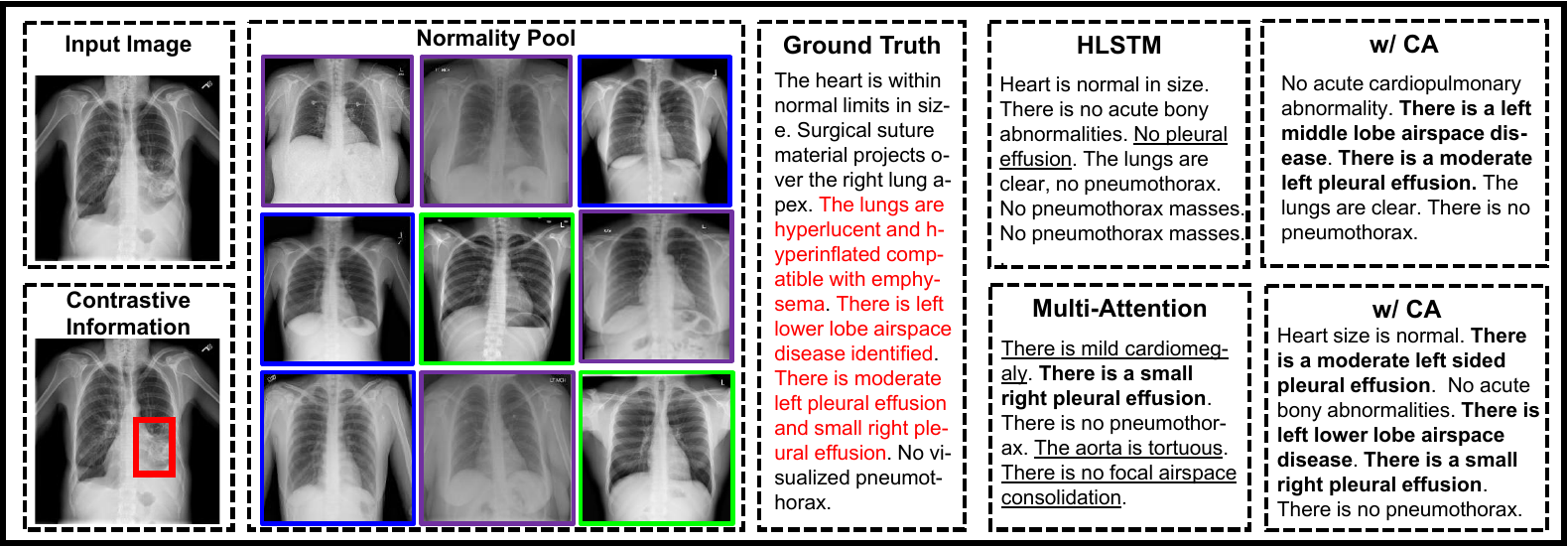}
\caption{Examples of the generated reports and the visualization of the Contrastive Attention (CA). Please view in color. CA model can capture the abnormal region ({\color{red} Red} bounding box) by contrasting the input image with normal images.  Besides, our Aggregate Attention in the CA model can find the closest normal images ({\color{blue} Blue} and {\color{green} Green} boxes visualized from two different attention weights (see Eq.~(\ref{eqn:n}))) and filter out the noisy images ({\color{purple} Purple} boxes).  The {\color{red} Red} colored text denotes the abnormal descriptions in the ground truth report. \underline{Underlined} text denotes the generated wrong sentences. \textbf{Bold} text denotes the generated true abnormalities. As we can see, the abnormal region and the abnormal descriptions generated by our method show significant alignment with ground truth reports.}
\label{fig:example}
\end{figure*}

\smallskip\noindent\textbf{Effect of $n$} \
In this section, we analyze the effect of $n$ in our Aggregate Attention (see Eq.~(\ref{eqn:n})).
Table~\ref{tab:ablation}(f-j) shows that when $n$ is smaller than 6, the performance will increase with as $n$ increases.
The reason may be that repeating the Aggregate Attention $n$ times with different learnable attention weights can encourage the model to identify the closest normal images from $n$ aspects, e.g., organs or tissues, i.e., if an aspect of the normal image is similar to the input image, it will be identified as the closest normal image.
In this way, the Aggregate Attention can capture accurate and robust closest normal images from $n$ aspects.
To verify this, we randomly visualize two attention weights of Aggregate Attention in Figure~\ref{fig:example}.
As we can see, the Aggregate Attention can indeed identify the closest normal images from multiple aspects, e.g., `Bone/Clavicle' (Blue boxes) and `Right Lung' (Green boxes), which proves our arguments.
Moreover, larger $n$ will bring noise, i.e., normal images that are not similar to the current input image, to the model and thus impair the performance.

\subsection{Qualitative Analysis}
In this section, we show the reports generated by the Baseline models, i.e., HLSTM and Multi-Attention, and the ``Baseline w/ CA'' models, and the visualization in Figure~\ref{fig:example} to analyze the strength of our Contrastive Attention model intuitively.
As we can see, for the HLSTM, it tends to produce repeated findings and normal findings, which results from the overwhelming normal findings in the dataset, i.e., data deviation \cite{Shin2016Learning}.
For the Multi-Attention, with the help of attention mechanism, it can describe abnormalities, but some abnormalities are incorrect (Underlined text).
The reason is that it is difficult for the Multi-Attention model to efficiently learn the medical expertise from the dataset with data deviation to correctly detect the abnormal regions.
Since our Contrastive Attention model can efficiently capture the suspicious abnormal regions by contrasting the input images and normal images and transfer such power to the downstream models and datasets, we can thus help multiple baseline models to detect and describe comprehensive and accurate abnormalities.
As a result, the ``Baseline w/ CA'' models can generate fluent and accurate reports supported with accurate abnormal descriptions, showing significant alignment with ground truth reports.

\section{Conclusion}
\label{sec:conclusion}
In this paper, we propose the Contrastive Attention model to capture abnormal regions by contrasting the input image and normal images for chest X-ray report generation.
The experiments on two public datasets demonstrate the effectiveness of our approach, which can be easily incorporated into existing models to boost their performance under most metrics.
The clinical efficacy scores and human evaluation further prove our arguments and the effectiveness of our approach in helping existing models capture and depict the abnormalities.
Specifically, we achieve the state-of-the-art results on the two datasets with the best human preference, which could better assist radiologists in clinical decision-making and reduce their workload.

In the future, there are two potential ways to improve the contrastive attention. 
First, it may be better to perform the contrastive attention in path feature rather than global feature. 
Second, it may be better to utilize multiple feature maps from different convolutional layers rather than just the feature maps of last convolutional layer.

\section*{Acknowledgments}
This work is partly supported by Beijing Academy of Artificial Intelligence (BAAI).
We would like to sincerely thank the clinicians (Xiaoxia Xie, Jing Zhang, etc.) of the Harbin Chest Hospital in China for providing the human evaluation.
We also sincerely thank all the anonymous reviewers and chairs for their constructive comments and suggestions.
Xu Sun and Ping Zhang are the corresponding authors of this paper.

\section*{Ethical Considerations}

In this work, we focus on helping several existing chest X-ray report generation systems better capture and describe the abnormalities.
To this end, we provide a detailed human evaluation in Table~\ref{tab:human_evaluation} (Section \ref{sec:human_evaluation}) and an automatic evaluation in terms of clinical efficacy metrics in Table~\ref{tab:automatic_CE} (Section \ref{sec:automatic_CE}) to know (1) on what fraction of images with abnormalities did the system not mention the abnormality and (2) on what fraction of images the system described abnormality that does not exist according to doctors.
The results show that our work can help existing systems generate more accurate descriptions for clinical abnormalities, improving the usefulness of existing systems in better \textbf{assisting} radiologists in clinical decision-makings and reducing their workload.
In particular, for radiologists, given a large amount of medical images, the systems can automatically generate medical reports, the radiologists only need to make revisions rather than write a new report from scratch.
This study uses the public MIMIC-CXR and IU-X-ray datasets.
All protected health information was de-identified.
De-identification was performed in compliance with Health Insurance Portability and Accountability Act (HIPAA) standards in order to facilitate public access to the datasets.
Deletion of protected health information (PHI) from structured data sources (e.g., database fields that provide patient name or date of birth) was straightforward. 
All necessary patient/participant consent has been obtained and the appropriate institutional forms have been archived.

\bibliographystyle{acl_natbib}
\bibliography{acl2021}

\end{document}